\documentclass[nohyperref]{article}

\usepackage{microtype}
\usepackage{graphicx}
\usepackage{subfigure}
\usepackage{booktabs} 

\usepackage{url}

\usepackage{breakurl}
\usepackage[breaklinks]{hyperref}

\usepackage[accepted]{icml2022}

\usepackage{amsmath}
\usepackage{amssymb}
\usepackage{mathtools}
\usepackage{amsthm}

\theoremstyle{plain}

\theoremstyle{definition}

\theoremstyle{remark}

\usepackage[textsize=tiny]{todonotes}

\icmltitlerunning{Human-AI Collaborative Decision-Making: Beyond Learning to Defer}

\begin{document}

\twocolumn[
\icmltitle{Human-AI Collaboration in Decision-Making: Beyond Learning to Defer}




\begin{icmlauthorlist}
\icmlauthor{Diogo Leitão}{fdz,ist,it}
\icmlauthor{Pedro Saleiro}{fdz}
\icmlauthor{Mário A. T. Figueiredo}{ist,it}
\icmlauthor{Pedro Bizarro}{fdz}
\end{icmlauthorlist}

\icmlaffiliation{fdz}{Feedzai}
\icmlaffiliation{ist}{Instituto Superior Técnico, Universidade de Lisboa}
\icmlaffiliation{it}{Instituto de Telecomunicações}

\icmlcorrespondingauthor{Diogo Leitão}{diogo.leitao@feedzai.com}

\icmlkeywords{Human-AI collaboration, Learning to defer, Fairness, Robustness, Selective Labels}

\vskip 0.3in
]



\printAffiliationsAndNotice{}  

\begin{abstract}
\textit{Human-AI collaboration} (HAIC) in decision-making aims to create synergistic teaming between human decision-makers and AI systems.
\textit{Learning to defer} (L2D) has been presented as a promising framework to determine who among humans and AI should make which decisions in order to optimize the performance and fairness of the combined system.
Nevertheless, L2D entails several often unfeasible requirements, such as the availability of predictions from humans for every instance or ground-truth labels that are independent from said humans.
Furthermore, neither L2D nor alternative approaches tackle fundamental issues of deploying HAIC systems in real-world settings, such as capacity management or dealing with dynamic environments.
In this paper, we aim to identify and review these and other limitations, pointing to where opportunities for future research in HAIC may lie.
\end{abstract}

\section{Introduction}
\label{introduction}
With the advent of machine learning (ML), artificial intelligence (AI) is now fast, scalable, and, in several applications, very accurate.
As a result, AI is increasingly present in decision-making processes previously conducted solely by humans, such as in healthcare \citep{esteva2017dermatologist}, or in criminal justice \citep{kirchner2016compas}.
The AI may take on a pure advisory role, providing the human decision-maker with a prediction, a score, or explanations.
However, in several applications, the volume of cases is so large that it is impractical or even economically unfeasible to solely rely on human decision-making.
This is the case in fraud detection \citep{yousefi2019fraud}, and credit scoring \citep{burrell2016opacity}.
Because AI models are able to process large volumes of information, they offer superior operational efficiency and statistical precision to that of humans, who are known to struggle in this regard \citep{meehl1954clinical}.
With that said, humans are capable of swiftly learning from few data points, and can leverage information that the AI cannot grasp, either because it cannot be converted to symbolic representation, or because it originates from accrued experience.
As such, often the ideal system consists neither of AI alone, nor of humans alone, but rather of a collaboration between the two.

A key challenge in \textit{human-AI collaboration} (HAIC) is to determine who makes which decisions.
An ideal HAIC system is capable of identifying the strengths and weaknesses of humans and AI in order to optimally assign instances to decision-makers.
This encompasses modelling not only predictive performance, but also \textbf{fairness}.
ML models are known to be capable of producing biased decisions against protected groups \citep{angwin2016bias, buolamwini2018gender}.
Humans, too, can be prejudiced, especially when information correlated with protected attributes is available to them.
HAIC systems have the opportunity to reduce unfairness by optimizing assignments to mitigate group-wise disparities.

The current state-of-the-art approach for performance- and fairness-aware assignments in HAIC is \textit{learning to defer} (L2D) \citep{madras2018predict, mozannar2020consistent, keswani2021multiple}.
L2D frames the assignment problem as a classification task, providing a supervised method to find the best decision-maker for each given instance.
Nevertheless, L2D exhibits structural limitations: it \textbf{requires predictions from every human decision-maker for every instance} in the training set, and trades off \textbf{robustness} for specialization.
Additionally, \textbf{capacity management} in teams of humans is disregarded: L2D aims to find the optimal assignment given an individual instance, instead of aiming to find the optimal set of assignments for a set of instances given a team of humans with limited capacity. 

Settings where decisions influence outcomes are also usually not considered.
A particularly ubiquitous problem is \textbf{selective labels} \citep{lakkaraju2017selective}.
For instance, in criminal justice, rearrest can only take place if bail is first granted; in credit scoring, failing to repay a loan can only happen if a loan is granted.
Lastly, a key issue still unaddressed in the literature is dealing with \textbf{dynamic environments}, which may entail addressing concept drift \citep{gama2014drift} and feedback loops \citep{dalvi2004adversarial, perdomo2020performative}.
To avoid performance degradation, HAIC systems must account for non-stationary factors, or at least be updated with new data.
However, L2D offers no solution to learn from partial, non-random data, and neither do alternative contributions.

In short, HAIC systems have the opportunity to optimize performance and fairness beyond that of individual contributors.
Although L2D provides a valid framework to this end, it entails several structural limitations that inhibit its adoption in real-world use-cases and disregards other key challenges.
The goal of this paper is to clearly define these challenges and limitations, while shedding light on possible avenues of future research to bring us closer to performant, fair, and practical human-AI collaboration.

\section{Learning to Defer}
\label{l2d}
The simplest approach to manage assignments in HAIC is to defer based on model confidence.
This approach is drawn from learning with a reject option, a framework first studied by \citet{chow1970reject}, where the goal is to optimize the performance of the non-rejected predictions by rejecting to predict in cases of high uncertainty.
The baseline for uncertainty estimation is to take the predicted probability score(s) as a measure of confidence \citep{hendrycks2016baseline}, while more sophisticated techniques range from ensemble methods \citep{gal2016dropout, lakshminarayanan2017ensembles} to auxiliary models \citep{corbiere2019auxiliary}.
In HAIC systems, these uncertainty estimation techniques can be leveraged to defer low model confidence cases to humans.

\citet{madras2018predict} introduced learning to defer (L2D) as an improvement upon confidence-based deferral in the context of HAIC systems with a single human expert.
The authors argue that confidence-based deferral is suboptimal, as it fails to consider the downstream human decision-maker.
In some instances of high model uncertainty, they may be just as inaccurate as the model; as such, it may be preferable to defer to them other lower uncertainty instances where they are able to outperform the model.
To model these differences, \citet{madras2018predict} expand upon the framework of learning with a reject option.
In particular, they adapt the work of \citet{cortes2016rejection}, who incorporated the reject option into the learning algorithm, allowing the classifier to reject to predict, incurring instead in a constant, label-independent cost.
To account for the performance of the human decision-maker, L2D adds a variable cost of deferral to this setup:
\begin{alignat*}{2}
& \mathcal{L}_{\textrm{\textit{system}}}(\hat{Y}_M, \hat{Y}_H, s) = \sum_i [&&(1-s_i) \mathcal{L}_{\textrm{\textit{CE}}}(Y_i, \hat{Y}_{M,i}) \\
& &&+ s_i \mathcal{L}_{\textrm{\textit{0-1}}}(Y_i, \hat{Y}_{H,i}) + s_i \gamma_{\textrm{\textit{defer}}}]
\end{alignat*}

The system loss depends on the choice to defer ($s=1$) or not to defer ($s=0$) to the human decision-maker $H$.
If the human decision-maker is chosen, the system loss is the 0-1 loss between ground-truth label $Y$ and human prediction $\hat{Y}_H$, plus a constant penalty term $\gamma_{\textrm{\textit{defer}}}$.
Conversely, if the decision is not deferred, the system loss corresponds to the cross-entropy loss of model $M$'s prediction $\hat{Y}_M$.

\citet{madras2018predict} propose a learning scheme consisting of two separate neural networks that are jointly trained to minimize the system loss.
The main model focuses on the classification task at hand; the deferral model decides whether to assign the decision to the model or to the human decision-maker.
During training, the output of the deferral model is taken as a Bernoulli probability, with both networks backpropagating on the expected value of the system loss.
Consequently, the deferral model learns to choose the decision-maker with best expected performance, and the main classifier is allowed to specialize on the cases that are not deferred, disregarding the rest.

An additional concern of \citet{madras2018predict} is that both the human decision-maker and the ML model may be biased against certain protected groups.
To allow the L2D system to mitigate these biases, the authors propose a regularized loss with an additional penalty term for error-rate disparities between protected groups.
We will further develop on the topic of fairness in HAIC in Section~\ref{fairness}.

Others authors have since joined the effort of expanding and improving learning to defer.
\citet{mozannar2020consistent} generalize L2D to multi-class settings and introduce a consistent surrogate loss to be used in a single classifier capable of deferral.
\citet{verma2022calibrated} elaborate upon their work, proposing a loss calibrated for human correctness.
\citet{keswani2021multiple} expand L2D to model multiple human decision-makers.
Lastly, some authors propose different methodologies \citep{wilder2020complement, gao2021bandit}, or different end-goals \citep{bansal2021teammate}.
In the next section, we will shed light on the limitations of L2D, as well as unaddressed challenges by this and alternative frameworks.

\section{Challenges and Limitations}
To help illustrate the shortcomings of L2D, we will take the domain of online merchant fraud detection as a running example.
In this use-case, ML models and human fraud analysts work in tandem to identify and stop fraud in streams of thousands of daily purchases.
As such, the AI is used to process the majority of purchases, while humans, due to limited capacity, only intervene on a small fraction (e.g. 5\%) of the most difficult instances.


\subsection{Learning from data without human predictions}
\label{complete}
Learning to defer entails a fundamental, structural limitation: it requires predictions from human decision-makers for every training instance.
This implies an additional effort that goes much beyond what is strictly necessary in regular operations.
One cannot expect a team of fraud analysts hired to intervene only on the most difficult cases to be able to review all the instances necessary to train an ML model.

\citet{mozannar2020consistent} consider this problem, and suggest an imputation scheme where the behavior of the human decision-makers is modelled from the complete data and replicated on the incomplete data.
However, that method fails when the complete and incomplete data are not identically distributed, as the basis for generalization falters.
This will be the case whenever the collected data stems from a non-random assignment system.
In real-world use-cases, this is to be expected, as aforementioned techniques such as deferring low-confidence instances to the human decision-makers already bring substantial performance gains.
In fraud detection, analysts do not review random cases; they review cases where the ML model is less confident.

As such, L2D cannot be implemented as an improvement to an existing deferral system.
In fact, if a L2D system needs to be updated to keep up with changing trends in the data, the framework cannot be reused to perform this update, as non-deferred instances will lack a human prediction, and will be distributed differently from deferred instances.
We will revisit this point in Section~\ref{nonstat}, elaborating on why model updating will, in fact, be necessary in most settings.

Future research should focus on being able to learn from data with and without human predictions, as that is its natural state in HAIC settings.
An interesting approach would be to resort to reinforcement learning techniques, where partial and biased information is the norm.
Choices of decision-maker could be framed as actions, with the environment being the features and any other relevant information.

\subsection{Joint learning: losing robustness and advisory}
One of the proposed benefits of L2D is that, by jointly training the main classifier with the deferral model, the former is allowed to specialize on non-deferred instances, in detriment of those that are deferred \citep{madras2018predict, mozannar2020consistent}.
However, by design, specialization renders the main classifier useless in the instances likely to be deferred, as gradients are stopped from backpropagating into it.
As such, \textbf{learning to defer is unsuitable in domains where the AI advises the human decision-makers}.
In such use-cases, the ML model may provide the human decision-maker with a tentative prediction, a score, or a set of explanations, that they will then combine with their own accrued experience and analysis of the instance at hand.
In fraud detection, the ML model score is one of the pieces of information considered by analysts when making a prediction \citep{jesus2021xai}.
This would not be possible if the classifier specialized away from the deferred instances.

Furthermore, \textbf{specialization makes the system brittle}: by trading off generalization for specialization, the AI is not robust to post-deployment changes on human capacity for review.
If fraud analysts become temporarily unavailable, the AI will not be capable of deciding on a fraction of instances, seriously harming performance.

A promising avenue of research could be to develop an alternative approach to L2D that follows a sequential learning setup instead of joint learning.
This option is briefly considered by \citet{madras2018predict}, when the authors suggest stopping system error from backpropagating to the main classifier; however, results of this alternative setup are not shown, and it has not been considered in posterior contributions.
A HAIC assignment system trained sequentially would solely focus on finding the best assignment policy given the pre-trained classifier and the set of human decision-makers.
This approach would not only avoid the aforementioned drawbacks of altering the main classifier, but also eliminate the need for human predictions for every instance, which, as mentioned in Section~\ref{complete}, is often an unfeasible demand.

\subsection{Deferral to multiple decision-makers}
\citet{madras2018predict} introduce learning to defer in the context of \textbf{deferral to a single human decision-maker}.
However, in several applications, teams may be composed of several people, and they may have different biases and types of expertise.
For example, fraud analysts may be specialized in specific fraud patterns.

\citet{keswani2021multiple} propose a L2D method that accounts for the diversity of human decision-makers, modelling each available expert individually.
However, to this end, they require not only human predictions for every instance --- as in single-human L2D --- but also predictions from every considered decision-maker.
This data will rarely be available in real-world use-cases, as it is inefficient for individuals in teams of human experts to review each other's decisions.
If every fraud analyst in a team with $n$ elements starts reviewing every deferred instance, the capacity for reviews decreases by a factor of $n$, harming the system's performance.
\citet{keswani2021multiple} posit that missing predictions may be imputed; however, the robustness of that method is not discussed.
If the data reflects non-random assignments, then generalization may not be possible.

Future work should focus on modelling several decision-makers without requiring concurrent predictions, as those will rarely be available without additional effort.
As mentioned in Section~\ref{complete}, this could be accomplished with reinforcement learning techniques, where partial information is the norm.


\subsection{Capacity management}
\label{capacity}
Learning to defer frames deferral as a classification task: given a specific instance, the goal is to find which decision-maker is expected to be the most accurate and less biased.
However, in the presence of capacity constraints, the best decision-maker may not be the optimal choice.
For example, if a fraud analyst is universally better than the rest of the team, then, ideally, they would decide on only the hardest cases where others are most likely to err.

Current L2D methods would simply assign them to every instance, disregarding capacity constraints.
While this may be acceptable in use-cases where the hiring process is fully flexible, such as with Amazon Mechanical Turk, in human-AI teams with stricter arrangements, such extreme assignments are unfeasible.
As a result, for an optimal deferral solution, capacity must be managed explicitly.
We believe this to be a blind spot and an opportunity for future research.

\subsection{Selective labels}
Human-AI decision-making teams often operate in environments where actions can influence outcomes.
In fraud detection, a purchase may only be found to be fraudulent if it is not declined.
\citet{lakkaraju2017selective} refer to this as the selective labels problem.
L2D is incapable of dealing with selective labels, as it requires ground-truth labels to evaluate human decision-makers on.

\citet{gao2021bandit} propose an alternative method anchored not on predictions and ground-truth labels, but rather actions and rewards.
For instance, in fraud detection, the reward of past declined purchases is always zero, regardless of the (unobservable) ground-truth label, as no revenue or loss occurs.
On the other hand, predicted negatives either have a negative reward --- the cost of accepting a fraudulent purchase ---, or a positive reward --- the revenue of a legitimate purchase.
\cite{gao2021bandit} propose using this bandit feedback to learn both a decision-making model and a routing model, leveraging counterfactual risk minimization \citep{swaminathan2015counterfactual}.
Despite modelling rewards instead of labels, their approach still requires human predictions for every training instance.

Alternatively, selective labels can also be directly addressed.
Imputation techniques may be used, leveraging expert consensus when available \citep{dearteaga2018selective, keswani2022closed}.
Furthermore, selective labels do not completely eliminate relevant signal, as predicted negatives will still result in either true or false negatives.
Consequently, performance can be compared on the basis of the trade-off between predicted positive prevalence and false negative prevalence \citep{lakkaraju2017selective}.
A HAIC system based on these signals could still be beneficial, without requiring additional data.

\subsection{Fairness}
\label{fairness}
As mentioned in Section~\ref{introduction}, both humans and ML models are capable of bias against protected groups \citep{angwin2016bias, buolamwini2018gender}.
When introducing learning to defer, \citet{madras2018predict} were keen on building a framework that optimizes not only for performance, but also for fairness.
Considering the ML model's biases, as well as those of humans, allows the deferring system to mitigate existing biases.
On the contrary, introducing customized deferral systems without considering fairness may result in aggravating existing biases, as demonstrated by \citet{jones2020selective} in the context of learning with rejection.

Fairness in HAIC represents not only an opportunity, but also a threat, and should thus always be considered.
While most contributions on L2D do consider fairness \citep{madras2018predict, keswani2021multiple}, other contributions inside and outside L2D do not \citep{mozannar2020consistent, gao2021bandit, bansal2021teammate}.
With AI being increasingly used to support decision-making in key societal sectors, future research should aim to maintain fairness at the forefront of human-AI collaboration.

\subsection{Dynamic environments}
\label{nonstat}
Lastly, a key unaddressed issue is dealing with dynamic environments.
The underlying data distribution may change due to new trends, resulting in concept drift \citep{gama2014drift}.
Furthermore, in performative prediction settings \citep{perdomo2020performative}, the actions of the system may influence the environment.
In adversarial classification settings \citep{dalvi2004adversarial}, it may even elicit a response (e.g. fraudsters changing strategies as a response to fraud detection systems).
The system's decision-makers may also change their behavior, as they learn new concepts, update beliefs, or are influenced by exogenous factors.
HAIC systems must account for all these non-stationary factors, or at least be updated with new data to avoid performance degradation.

L2D does not endogenously model any of the aforementioned non-stationarity factors, and it cannot be updated with new data naturally obtained from the HAIC environment (see Section~\ref{complete}).
Future research should focus on developing HAIC systems that do not have this limitation.


\section{Conclusion}
In this paper, we shed light on the structural limitations and unaddressed challenges of the state-of-the-art method for managing assignments in human-AI collaborative decision-making --- learning to defer.
Our goal is to motivate research moving towards a holistic human-AI collaboration system that learns to optimize performance and fairness from the available data, while managing existing capacity constraints, and keeping up with dynamic environments.

\newpage
\Urlmuskip=0mu plus 1mu\relax
\bibliographystyle{icml2022}
\bibliography{references}
\end{document}